\documentclass{article} % For LaTeX2e
\usepackage{iclr2027_conference,times}

\usepackage{amsmath,amsfonts,bm}

\def\eqref#1{equation~\ref{#1}}
\def\1{\bm{1}}

\DeclareMathAlphabet{\mathsfit}{\encodingdefault}{\sfdefault}{m}{sl}
\SetMathAlphabet{\mathsfit}{bold}{\encodingdefault}{\sfdefault}{bx}{n}

\usepackage{hyperref}
\hypersetup{
  colorlinks=true,
  linkcolor=blue,
  citecolor=blue,
  urlcolor=blue,
}
\usepackage{url}
\usepackage{graphicx}
\usepackage{subcaption}
\usepackage{pifont}
\usepackage{booktabs}
\usepackage[normalem]{ulem}
\usepackage{enumitem}

\title{From Representation to Readout: Mechanistic Analysis of Membership Signals in Language Models}

\author{
Wenxi Dai$^{1}$, Jingyu Hu$^{2}$, Yanting Yang$^{1}$,
\textbf{Mengnan Du}$^{3}$\\
{\small $^{1}$Rutgers University \quad
$^{2}$University of Bristol}\quad
{\small $^{3}$The Chinese University of Hong Kong, Shenzhen}\\
{\small \texttt{daiwenxi3@gmail.com},\,\, \texttt{mengnandu@cuhk.edu.cn}}
}

\iclrfinalcopy % Uncomment for camera-ready version, but NOT for submission.
\begin{document}

\maketitle

\begin{abstract}

%Activation patching further provides causal evidence that this membership signal emerges in the middle-to-late layers.
%150-180 words
%Membership inference aims to determine whether a particular example was included in a model’s training data. 我们当然可以笼统地称这种membership signals为memorization，但是这个说法并不精确，但是it remains unclear how such signals are presented internally and how they affect model behavior. 所以？We present a mechanistic study of membership signals in large language models。

%This raises a mechanistic question: how are membership signals represented internally, and how does the model access them to produce behavior? 

%Across training paradigms, we find that changes in membership-related behavior are not accompanied by comparable increases in decodable membership information, but are instead primarily reflected in changes in readout.

Membership inference aims to determine whether a particular example was included in a model’s training data. Such membership signals are often broadly described as memorization, while the underlying mechanisms can involve both how information is represented internally and how it is accessed by the model to produce behaviors. We therefore present a mechanistic study of membership signals in large language models (LLMs). We introduce Decodable Information Score, Used Information Score, and Readout Accessibility Gap (DUG), a set of complementary measures that separately quantify membership information decodable from internal representations and the extent to which it is exposed through the model’s readout. Across supervised fine-tuning (SFT) and reinforcement learning (RL)-based membership inference methods, membership-specific effects appear in both the amount of decodable information and readout under SFT, while under RL-based membership inference methods they appear primarily in readout. Activation patching further localizes this membership-specific effect to intermediate layers. To examine whether these findings extend to a more general RL setting, we further study reinforcement learning with verifiable rewards (RLVR) contamination, where the training reward is independent of the reference solution. In this setting, we find that behavioral changes can emerge without a corresponding increase in likelihood-based membership inference signals.

\end{abstract}

\section{Introduction}
Membership inference attacks (MIAs) aim to determine whether a given example belongs to a model’s training data. They have been widely studied as a way to characterize and audit a model's exposure to training data. We refer to included examples as members and excluded examples as non-members \citep{shokri2017membershipinferenceattacksmachine,duan2024membershipinferenceattackswork}. Prior work uses signals from the behavior of large language models (LLMs), such as loss, perplexity, and model outputs, to infer membership \citep{xie-etal-2025-memorization, carlini2021extractingtrainingdatalarge}. More recent approaches further modify the target model weights to bring out membership-related behaviors, typically using reinforcement learning (RL) or supervised fine-tuning (SFT) to elicit membership signals \citep{yin2026learningdetectlanguagemodel, fu2024miatuneradaptinglargelanguage}. Such signals are often broadly referred to as memorization.

Interpretability studies of memorization have provided evidence that information associated with memorized content can manifest internally, and have shown that such representations can be actively steered at inference time to affect model behavior \citep{stoehr2024localizingparagraphmemorizationlanguage}. Much of this work studies memorization through the model's ability to reproduce training content. Some studies group examples by their reproducibility and analyze the resulting differences \citep{chen2026comparativeanalysisllmmemorization, wang2025generalization}. Others manipulate training exposure and examine the resulting changes in model behavior \citep{lee-etal-2022-deduplicating}. These analyses primarily ask how well the model reproduces a given example, rather than directly comparing members and non-members to characterize how training membership is reflected in the model's internal representations. In particular, it remains unclear how these interventions alter how models expose information associated with their training data.

Motivated by this gap, we ask: \emph{\uline{How do training interventions shape membership signals in language models? --by changing the information encoded in internal representations, by changing how that information is read out, or both?}} To address this question, we introduce the \textbf{D}ecodable Information Score, \textbf{U}sed Information Score, and Readout Accessibility \textbf{G}ap, collectively termed \textbf{DUG}, a layer-wise set of complementary measures that quantify membership information in internal representations and its accessibility through the model’s readout. We combine these measures with a difference-in-differences analysis to isolate membership-specific changes induced by training. We then apply this framework to SFT and RL-based MIA methods, further using activation patching to causally localize where the induced membership-specific effect forms, and examine whether these findings extend to a more general RLVR contamination setting. 

To be specific, our main contributions can be listed as follows:

\begin{itemize}[leftmargin=10pt, topsep=-2pt, itemsep=1pt, partopsep=1pt, parsep=1pt]
    %\item We propose \textbf{DUG}, together with a difference-in-differences analysis to decompose membership-specific changes in decodable information and model readout, and to quantify how training changes each component across layers.
    \item We propose \textbf{DUG}, together with a difference-in-differences analysis that separates membership-specific changes from those shared with non-members, decomposes them into decodable information and model readout, and quantifies both across layers.

    \item Across controlled SFT and RL-based membership inference methods, we find that membership-specific changes under SFT emerge in both decodable information and readout, whereas RL-based MIA methods primarily affect the readout of membership-specific information (Section~\ref{sec:dug_findings}). We further use activation patching to causally localize this effect across layers (Section~\ref{sec:patching}).

    \item We show, in a relatively general RLVR contamination setting where the reward is independent of the reference solution, that behavioral membership effects can emerge without a corresponding increase in likelihood-based MIA performance, thereby revealing a dissociation between behavioral and likelihood-based membership inference (Section~\ref{sec:rlvr_sec}).

\end{itemize}

\section{Related Work}

\subsection{Membership Inference Attack in LLMs}

MIA aims to determine whether a given data sample was included in a model’s training data by exploiting differences in the model’s behavior on member and non-member samples \citep{shokri2017membershipinferenceattacksmachine, duan2024membershipinferenceattackswork}. Prior work has explored methods that actively induce models to expose membership signals through targeted interventions, including both training-based and training-free approaches. \cite{rl-data-contamination} argue that RL creates a detectable membership signal through policy collapse, as rewarded contaminated samples show more stable and less diverse reasoning when probed for membership. 
Similarly, \cite{zhang2026detectingrlvrtrainingdata} link RLVR-induced policy collapse to membership inference, showing that membership signals become increasingly detectable as RLVR training progresses. 
\cite{yin2026learningdetectlanguagemodel} directly optimize models via RLVR for membership detection to elicit latent membership signals. 
These methods suggest that training can make membership signals more pronounced~\citep{liu2026auditingdatamembershipreinforcement, gwak2026laralayerwiserepresentationanalysis}. Controlled-contamination studies have also considered settings where only a small subset of evaluation data is introduced into training \citep{magar2022datacontaminationmemorizationexploitation, shi2024detectingpretrainingdatalarge, oren2024proving}. However, these works primarily evaluate signal detectability, while how training changes their representation and readout of remains unexplored.

\subsection{Interpretability of Memorization in LLMs}

Memorization in language models has been studied primarily through behavioral tests of whether a model can reproduce or extract training data. Existing works have quantified memorization using likelihood- or perplexity-based measures \citep{xie-etal-2025-memorization, carlini2021extractingtrainingdatalarge}, and used membership inference to assess whether training examples exhibit distinguishable behavior from non-members \citep{yan2026spuriousrewardsparadoxmechanistically, duan2024membershipinferenceattackswork, tao2025tokenlevelinformiastrongermembership}. %Recent work has further examined membership signals at the token level, enabling fine-grained localization within individual sequences \citep{}.

From a white-box perspective, prior work has investigated the internal mechanisms underlying what is broadly characterized as memorization, using mechanistic interpretability tools such as circuit discovery and activation steering to identify representations involved in initiating and maintaining memorized behavior \citep{makhija-etal-2026-neural, prabhakar-etal-2024-deciphering}. Related work on representation editing further shows that such internal representations can be intervened on to amplify or suppress specific behaviors \citep{stolfo2025improvinginstructionfollowinglanguagemodels}. These studies provide evidence that memorization can be linked to identifiable and manipulable internal representations. However, it remains unclear how the membership signals detected by MIA relate to these internal representations, and whether training primarily changes the information stored in the representation or how that information is read out.

%In this sense, data contamination can manifest as memorization when information from the training data becomes preferentially accessible to the model.

\section{Preliminary: Is Membership Linearly Decodable?}

\label{sec:prelim}

Before investigating how further training affects membership signals, we first study whether these signals are already encoded in publicly released models prior to any additional training, in a binary setting where members come from the model's own pre-training or post-training data. Specifically, we train linear probes on the residual-stream representations at each layer to assess the linear decodability of membership status from internal model representations. %The evaluation covers LLM-MIA benchmarks spanning both pre-training and post-training detection. 
For pre-training membership, we evaluate Qwen2-7B on WikiMIA-2024-Hard~\citep{hallinan2026surprisingeffectivenessmembershipinference}, where members and non-members are separated by a temporal cutoff. 

For post-training membership, we examine the open Tulu3 model on the Aya \citep{singh-etal-2024-aya} and WildChat \citep{zhao2024wildchat1mchatgptinteraction} subsets of its post-training mix, with members included in the mix and non-members held out.

\textbf{Baselines and Controls.} For each dataset, we compare our probe AUROC with a blind baseline that predicts membership from TF-IDF and surface text statistics alone without access to the model's representations \citep{das2025blindbaselinesbeatmembership}. We also retrain the probes after shuffling membership labels as a control. A probe is considered informative only if it outperforms the blind baseline while the shuffled control remains at chance level as expected. %Detailed analyses and further results are provided in Appendix~\ref{app:prelim}.

\begin{figure}[t]
    \centering
    \includegraphics[width=1\linewidth]{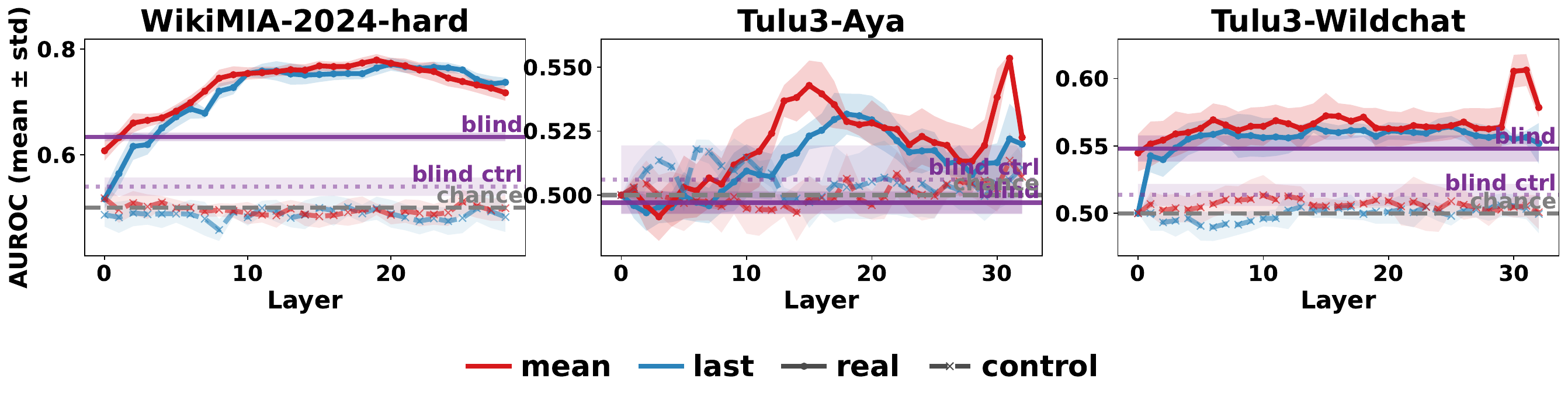}
    \caption{Layer-wise probe AUROC for membership prediction on three benchmarks.}
    \label{fig:preliminary1}
\end{figure}

%whose members enter the model during post-training,
\textbf{Findings.} Figure~\ref{fig:preliminary1} reports per-layer probe AUROC on the base model, measuring whether a linear probe can distinguish member from non-member sequences. 
On WikiMIA-2024-hard, probe AUROC rises sharply in early layers and peaks at 0.78 around layer 20, substantially exceeding the blind baseline. 
On Tulu3-Aya and WildChat, probe AUROC reaches 0.543 and 0.57, respectively, in the middle layers, and rises again to 0.553 and 0.605 at the second-to-last layer before dropping at the final layer. 
The late-layer pattern aligns with prior findings of output collapse at the final layer and richer features at the second-to-last layer \citep{belrose2025elicitinglatentpredictionstransformers,skean2025layer}.

These observations show that membership signals are linearly decodable from representations, though their strength and layerwise patterns differ across the two settings. However, binary probing alone does not reveal 
whether these differences arise from the encoded information, its readout, or both. Moreover, we include binary probing for the models studied in Section 5 in App.~\ref{app:binary-probe}, which provides limited resolution of the observed membership effects. This motivates our measures for disentangling these two aspects by measuring information at the level of individual target tokens.

\section{The DUG Framework}

In the previous section, we showed that membership is linearly decodable from model representations. Here, we move beyond binary membership classification to examine the information underlying this signal. We focus on the target tokens predicted by the model, using their target-predictive information to characterize membership rather than a membership label directly. We quantify how much membership-related information is decodable from the representations and how much is exposed through the model’s readout at each layer. Their difference captures information that is encoded in the representations but not exposed by the readout.

\subsection{Problem Statement}

We study frozen language models obtained by training a base model $\theta_0$. Membership is determined by a training set $\mathcal{T}$: evaluation questions that appear in $\mathcal{T}$ form the members $\mathcal{M}$, and the rest form the non-members $\mathcal{N}$, a partition shared by all models. A trained model $\theta_a$ is obtained by training the base model $\theta_0$ on $\mathcal{T}$ under a specified training procedure, such as SFT or an RL-based MIA method. We denote the corresponding training procedure by $a$.

Each question $p$ consists of a context $c_p$ and a target sequence $y_p^\star$. We teacher-force each model on $c_p \Vert y_p^\star$ and pair every target token with the residual-stream representation at the preceding position, writing the $i$-th pair as $(h_i^{(\ell)}, y_i^\star)$ with $h_i^{(\ell)} \in \mathbb{R}^d$ at layer $\ell$ and $y_i^\star \in \mathcal{V}$, the vocabulary. Let $\mathcal{I}_p$ denote the pairs from question $p$. For cross-fitting, we split the pairs into $K$ folds grouped by question, keeping each question's tokens together, with training and evaluation splits $\mathcal{S}_{\mathrm{tr}}$ and $\mathcal{S}_{\mathrm{ev}}$.

\subsection{Proposed Framework}

\paragraph{Motivation.} We seek to quantify how much information about the ground-truth next token $y_i^\star$ is decodable from the residual representation $h_i^{(\ell)}$. Since this information is not directly measurable,  we compare a probe trained on $h_i^{(\ell)}$ against a null baseline that has no access to it, and define their difference
in cross-entropy as the \textbf{Decodable Information Score}. We introduce this concept to get a sense of how much target-relevant information there is to begin with, and how it changes with depth.

\paragraph{Null Baseline.} For each fold, we define an intercept-only classifier $q_b = \operatorname{softmax}(b)$ with $b\in\mathbb{R}^{|\mathcal{V}|}$, constant across positions. Fitting $b$ on the training split $\mathcal{S}_{\mathrm{tr}}$ recovers the
empirical distribution of $y_i^\star$ over that split, so the resulting loss on the
evaluation split,
% $\mathcal{L}_{\mathrm{null}}(y_i^\star) = -\log q_b(y_i^\star), i \in \mathcal{S}_{\mathrm{ev}}.$
\begin{equation}
\mathcal{L}_{\mathrm{null}}(y_i^\star) = -\log q_b(y_i^\star),
\qquad i \in \mathcal{S}_{\mathrm{ev}}.
\end{equation}
is what remains predictable from the marginal label distribution alone. It carries no dependence on $\ell$, and we reuse it for another quantity defined below as well.

\textbf{Decodable Information.} We fit a linear probe $q_\phi(y \mid h) = \operatorname{softmax}(Wh + \beta)$ at each layer of  the model, and use $\mathcal{L}_\phi^{(\ell)}(y_i^\star) = -\log q_\phi(y_i^\star \mid h_i^{(\ell)})$
for its cross-entropy loss on the target token. 
We denote the decodable information score at layer $\ell$ as $D(\ell)$, and define it as the average reduction in cross-entropy relative to the null baseline on the same folds:

\begin{equation}
D(\ell) \;=\; \frac{1}{|\mathcal{S}_{\mathrm{ev}}|}
\sum_{i \in \mathcal{S}_{\mathrm{ev}}}
\Big[ \mathcal{L}_{\mathrm{null}}(y_i^\star) - \mathcal{L}_\phi^{(\ell)}(y_i^\star) \Big].
\end{equation}

%We use $D_a(\ell)$ when the model needs to be made explicit. 

% Since we have $K$-fold data rotating, 
The $K$-fold rotation evaluates each position with a probe fitted on the remaining folds. $D(\ell)$ therefore reflects how much of $y^\star$ a linear
probe can recover from layer $\ell$ without having seen that position during
fitting, indicating whether the information has entered the representation.

\textbf{Used Information.} We denote used information score as $U(\ell)$ to quantify how much target-relevant information is exposed by the residual stream at layer $\ell$ through the model’s own unembedding $W_U$. 
% held frozen:
We hold $W_U$ frozen and read the residual stream through it:

\begin{equation}
\mathcal{L}_{\mathrm{lens}}^{(\ell)}(y_i^\star)
= -\log \operatorname{softmax}\!\big(W_U \, \mathrm{LN}_f(h_i^{(\ell)})\big)_{y_i^\star}.
\end{equation}

As Equation \ref{eq:U} shows, $U(\ell)$ is defined as the corresponding reduction in cross-entropy.
% , similar to $D(\ell)$,
We use the same null baseline and the same folds as for $D(\ell)$.
% computed against the same null baseline and the same folds as $D(\ell)$. 
Since no parameters are fitted, $U(\ell)$ reflects what layer $\ell$ already contributes to the model's own prediction. 

\begin{equation}
U(\ell) \;=\; \frac{1}{|\mathcal{S}_{\mathrm{ev}}|}
\sum_{i \in \mathcal{S}_{\mathrm{ev}}}
\Big[ \mathcal{L}_{\mathrm{null}}(y_i^\star) - \mathcal{L}_{\mathrm{lens}}^{(\ell)}(y_i^\star) \Big],
\label{eq:U}
\end{equation}

where, intuitively, $D(\ell)$ asks what is present in the residual stream, while $U(\ell)$ asks what the model puts to use. We use the logit lens rather than a tuned lens~\citep{belrose2025elicitinglatentpredictionstransformers}, because tuned-lens translators are trained to match the final-layer distribution and thus absorb computation from later layers, and their trainable parameters can themselves introduce member/non-member differences unrelated to the representation. Details in App.~\ref{app:tuned_lens}.

\paragraph{Readout Accessibility Gap.} 
We denote the Readout Accessibility Gap as $G(\ell)$ and define it as the difference of the two scores:

\begin{equation}
G(\ell) \;=\; D(\ell) - U(\ell)
\;=\; \frac{1}{|\mathcal{S}_{\mathrm{ev}}|}
\sum_{i \in \mathcal{S}_{\mathrm{ev}}}
\Big[ \mathcal{L}_{\mathrm{lens}}^{(\ell)}(y_i^\star) - \mathcal{L}_\phi^{(\ell)}(y_i^\star) \Big].
\end{equation}
%The null baseline cancels in this difference. A positive $G(\ell)$ means a linear probe recovers more about $y^\star$ from layer $\ell$ than the model's own unembedding does. The information is present in the residual stream but not yet promoted to the output distribution. At the final layer the logit lens coincides with the model's own output head, so $G(L) < 0$ compares a full-capacity head with a rank-limited probe and is expected by construction; $D$ should be read as a lower bound throughout.
The null baseline cancels in this difference. A positive $G(\ell)$ means that a linear probe recovers more about $y^\star$ from layer $\ell$ than the model’s own unembedding does, suggesting that information present in the residual stream is not yet fully expressed in the output distribution. This interpretation is most meaningful away from the final layer: at $L$, the logit lens uses the model’s own output head, whereas the probe remains rank-limited, so $G(L)<0$ is expected. Accordingly, $D(\ell)$ should be interpreted as a lower bound on the information available in the representation.

\paragraph{Difference-in-Differences.}

Beyond the information scores themselves, we ask whether training changes them differently for members and non-members, comparing each group's change in scores from the base to the trained model. 

Let $X(\ell, p; \theta)$ denote any of the scores above, averaged over the pairs $i \in \mathcal{I}_p$ of question $p$.
Training from $\theta_0$ to $\theta_a$ moves this score by $\Delta X_p(\ell) = X(\ell, p; \theta_a) - X(\ell, p; \theta_0)$. We refer to the gap between the member and non-member averages of $\Delta X_p(\ell)$ as the difference-in-differences (DiD):

\begin{equation}
\mathrm{DiD}[X](\ell) \;=\;
\frac{1}{|\mathcal{M}|}\sum_{p \in \mathcal{M}} \Delta X_p(\ell)
\;-\;
\frac{1}{|\mathcal{N}|}\sum_{p \in \mathcal{N}} \Delta X_p(\ell).
\label{eq:did}
\end{equation}

Here $\mathrm{DiD}[X]$ is computed for both $D$ and $U$. 
Since $\mathrm{DiD}[\cdot]$ is linear in $X$, the $\mathrm{DiD}$ for $G$ can be computed as $\mathrm{DiD}[G] = \mathrm{DiD}[D] - \mathrm{DiD}[U]$.  
A positive $\mathrm{DiD}[D]$ indicates that training made membership-relevant information more decodable from the representations, while a positive $\mathrm{DiD}[U]$ indicates that training amplified how strongly the readout uses that information.
%\textcolor{blue}{DiD[G] indicates xxxx}

\section{Experiments}

\subsection{Experimental Setup}

\paragraph{Design.}
\label{sec:design_dug_sec5}
To evaluate our framework across different training methods, we consider our primary setting as post-training, where we use AIME and Olympia Math~\citep{ai-mo2025olympiad} with Tulu2-7B~\citep{ivison2023camelschangingclimateenhancing} and introduce membership through controlled SFT. As a robustness check, we additionally evaluate a pre-training setting on WikiMIA-2024-Hard with Qwen2-7B~\citep{yang2024qwen2technicalreport}, where membership is determined by the temporal split of the corpus. For all datasets, members are randomly sampled. We consider the following training settings to characterize how membership manifests across different forms and degrees of training exposure. Details regarding experimental settings, datasets, and choices of models are listed in App.~\ref{app:dug_dataset}.

\ding{182} Base model, in which membership comes from pre-training naturally using the temporal split.

\ding{183} SFT, in which we vary the number of training epochs to control the degree of exposure.

\ding{184} RL-based MIA Methods, where we include the Active Data Reconstruction Attack (ADRA) and ADRA+ \citep{yin2026learningdetectlanguagemodel}. Unlike the preceding methods, which simulate the introduction of membership data during training, ADRA is an active MIA method that applies RL-based reconstruction to elicit membership information from the contaminated model. ADRA+ further incorporates a membership prior from the loss-based MIA method Min-K\%++ \citep{shi2024detectingpretrainingdatalarge}. We describe how these methods are initialized and trained in App.~\ref{app:explanation_RL}.

\subsection{Findings From DUG Framework}
\label{sec:dug_findings}

\begin{figure}[t]
    \centering
    \includegraphics[width=1\linewidth]{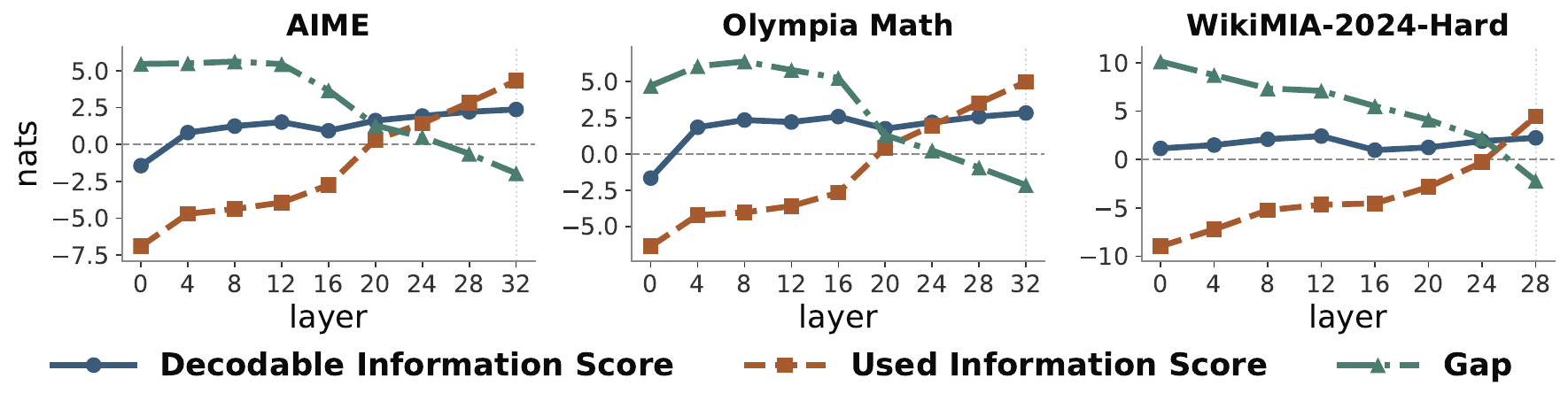}
    \caption{Results on base model.}
    \label{fig:DUG_base}
\end{figure}

\paragraph{Base Model Validation.} Figure~\ref{fig:DUG_base} shows the three information scores across layers on the base model $\theta_0$. $D(\ell)$ is already positive in early layers and increases with depth until it saturates. In contrast, $U(\ell)$ starts below the null baseline and rises with depth, suggesting that early residual representations are not yet well aligned with the unembedding space. Thus, the narrowing of $G(\ell)$ is primarily driven by $U(\ell)$ catching up with $D(\ell)$, consistent with later layers making encoded information accessible in the vocabulary space \citep{gurnee2026verbalizable}. Since early-layer $G(\ell)$ also reflects basis misalignment, we focus on its changes under training rather than its absolute value.

\begin{figure}[t]
    \centering
    \includegraphics[width=1\linewidth]{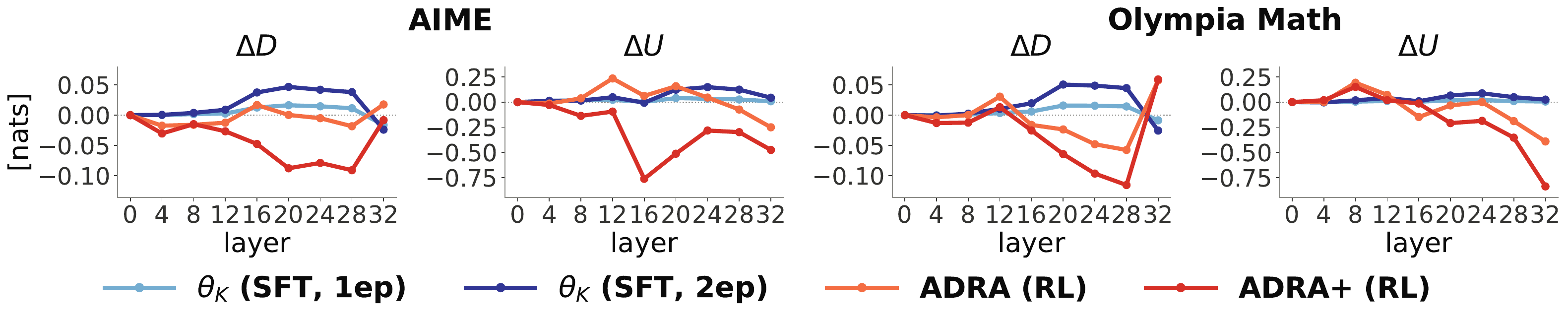}
    \caption{Change in Decodable Information and Used Information Score relative to $\theta_0$.}
    \label{fig:DUG_delta}
\end{figure}

\paragraph{Information Shifts Under Training.}
Figure~\ref{fig:DUG_delta} compares the absolute changes in $D$ and $U$ from the base model $\theta_0$ across the trained models ${\theta_a}$. Here, we focus on how training changes the overall representation and defer member-specific changes to the diff-in-diff analysis.

We find that training has a limited effect on decodable information in the residual stream, with $\Delta D$ from $-0.10$ to $+0.05$. In contrast, $\Delta U$  ranges from approximately $-0.75$ to $0.25$ nats, suggesting that training primarily affects how effectively the model reads out information already encoded in the residual stream. Notably, both RL-based MIA methods consistently reduce $\Delta U$ relative to the base model, reaching as low as $-0.83$ nats in later layers, indicating suppressed readout. In contrast, SFT induces only small positive changes in both $\Delta D$ and $\Delta U$. 
Within SFT, we further find that increasing the number of epochs leads to a gradual increase in $D$, consistent with prior findings that repeated exposure to the same training examples can increase memorization \citep{carlini2023quantifyingmemorizationneurallanguage}.

\begin{figure}[t]
    \centering
    \includegraphics[width=1\linewidth]{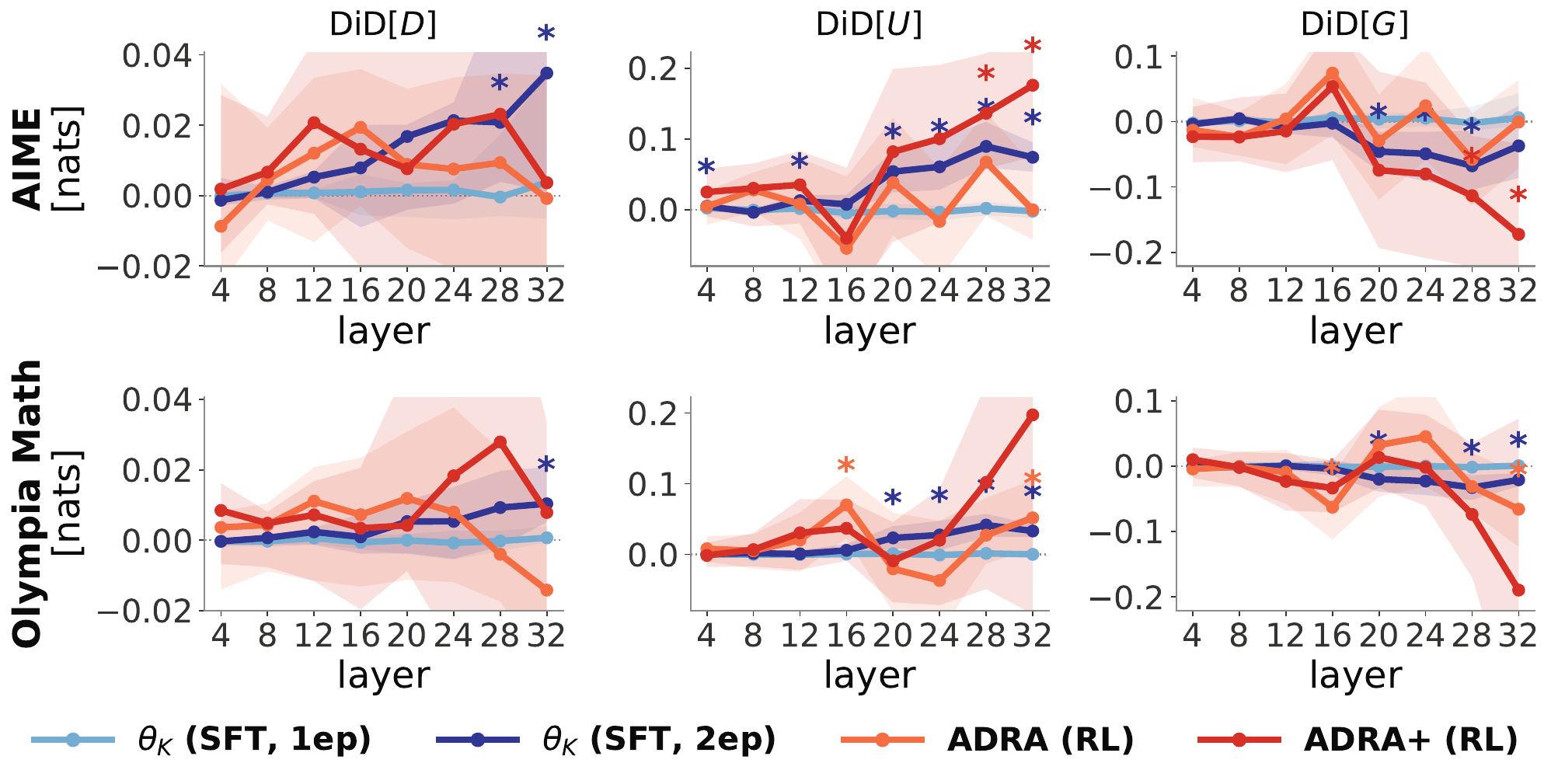}
    \caption{Diff-in-differences of membership-related changes in decodability ($D$), readout ($U$), and their gap ($G$) across layers. Shading shows 95\% CIs and * denotes significance.}
    \label{fig:DiD}
\end{figure}

\paragraph{Difference between Member and Non-member.} To examine whether training affects members and non-members differently, Figure~\ref{fig:DiD} reports the Difference-in-Differences (DiD) of $D$, $U$, and $G$ across layers. For $U$, we observe a clear member-specific effect. Previously, we have shown that RL-based MIA methods generally suppress the model’s readout. However, $\mathrm{DiD}[U]$ reveals that this suppression is substantially weaker for members than non-members. Specifically, ADRA+ yields positive $\mathrm{DiD}[U]$ values across most layers, reaching $+0.18$ nats at the final layer on AIME and $+0.20$ nats on Olympia Math. In comparison, $\mathrm{DiD}[D]$ does not exhibit a comparable member-specific shift. \emph{\uline{This suggests that RL-based MIA methods amplify the readout of member-specific information, while the decodable information itself shows no substantial member-specific change. }} This effect persists when $\mathrm{DiD}$ is computed relative to the SFT initialization, results in App.~\ref{app:sft1ep}.

\emph{\uline{Meanwhile, both $\mathrm{DiD}[D]$ and $\mathrm{DiD}[U]$ show larger differences when the number of epochs increases under SFT.}} This demonstrates a key distinction from RL-based MIA methods, which show a form of selective exemption for members in the aspect of readout suppression. One possible explanation is that the reward encourages the reconstruction trajectories of member problems to remain closer to $y^\star$, leading to less degradation in their readout. This differential suppression of readout may contribute to the membership signal observed by MIA.

\subsection{Case Study For RLVR}

\label{sec:rlvr_sec}
RL-based MIA methods take an attacker-centric perspective: given $x=(c,y^\star)$, they construct a reward function that encourages the model to reconstruct $y^\star$, eliciting latent membership signal. In practice, however, contamination takes a different form. RL in a post-training pipeline typically serves other objectives that have nothing to do with membership, e.g., improving mathematical or coding ability with verifiable rewards. In the course of pursuing these objectives, benchmark problems may end up in the RL prompt pool, thereby becoming members. 

To examine whether membership still leaves traces in the representations when RL objective is independent of $y^\star$, we present a case study on RLVR contamination. Because this is a controlled comparison, we deliberately keep the reward simple rather than engineering it for accuracy. Our main goal is to observe whether membership signal is detectable under this paradigm.

\paragraph{Setup.}We construct RLVR contamination on Qwen2-7B ($\theta_0$) using the MATH-500 dataset \citep{lightman2024lets}. 
Specifically, we mix 64 members into 640 GRPO training prompts (10\%) and hold out 64 nonmembers. 
% all with 
Both members and non-members are selected to satisfy $0 < \mathrm{pass@32} < 1$ to ensure their GRPO advantages are nonzero. 
RLVR checkpoints at steps 10 and 20 are exposure-matched to SFT epochs 1 and 2. Further details in App.~\ref{app:RLVR_setting}.

\paragraph{Objective.}

For each prompt $q$, we sample $G=32$ rollouts
$\hat{y}^{(i)} \sim \pi_\theta(\cdot\mid q)$. Each rollout receives a
verifiable reward:
\begin{equation}
r^{(i)}
=
\mathbf{1}\!\left[
\mathrm{verify}\!\left(\hat{y}^{(i)},a^\star\right)
\right] 
+ \lambda\,\mathbf{1}\!\left[
\hat{y}^{(i)} \text{ contains } \texttt{\textbackslash boxed\{\}}
\right],
\end{equation}

where $\mathrm{verify}$ tests symbolic equivalence with the gold answer $a^\star$
and the second term is a format bonus with $\lambda$ = 0.05. The reward uses only the final
answer and does not use the reference solution $y^\star$. We optimize with GRPO using
group-normalized advantages $A^{(i)}=(r^{(i)}-\mu_r)/\sigma_r$.

\paragraph{Accuracy Comparison and Training Validity.} 

We compare membership inference performance across the base model $\theta_0$, SFT with one and two epochs, and RLVR at different steps. We evaluate two likelihood-based MIA scores, Loss and Min-$k\%$, using AUROC. Table~\ref{tab:chat_mia} summarizes the results.

We can observe that membership leaves a far stronger trace under SFT. Loss AUROC rises from 0.491 at base to 0.661 after one epoch and 0.908 after two, and Min-k\% AUROC tracks it closely. Under RLVR the gain is far smaller, which we attribute to the training objective, in which the reward never references $y^\star$, so members are never directly fit. We can nonetheless provide evidence that RLVR trains effectively: across epochs, pool accuracy rises from 54.9\% to 71.8\%, with no signs of reward hacking, and the updates reach the members themselves (59 of 64 received at least one nonzero-advantage update). See Appendix~\ref{app:RLVR_valid} for details.

\begin{table}[t]
\centering
\resizebox{\columnwidth}{!}{%
\begin{tabular}{lcccc}
\toprule
\textbf{Model} & \textbf{Loss AUROC} & \textbf{$\Delta$AUROC}
& \textbf{Min-$k\%$ AUROC} & \textbf{$\Delta$AUROC} \\
\midrule
Base & 0.491 & -- & 0.492 & -- \\
SFT 1ep & 0.661 & +0.170 [+$0.102$, +$0.245$]
& 0.683 & +0.191 [+$0.109$, +$0.273$] \\
SFT 2ep & 0.908 & +0.417 [+$0.335$, +$0.508$]
& 0.916 & +0.424 [+$0.337$, +$0.522$] \\
RLVR (mid) & 0.493 & +0.002 [$-0.008$, +$0.013$]
& 0.490 & $-0.002$ [$-0.013$, +$0.008$] \\
RLVR (later) & 0.494 & +0.003 [$-0.009$, +$0.016$]
& 0.500 & $+0.008$ [$-0.007$, +$0.025$] \\
\bottomrule
\end{tabular}
}
\caption{Membership inference performance of Qwen2-7B on MATH-500 dataset. RLVR (mid) and RLVR (later) denote the intermediate and later RLVR checkpoints, respectively. $\Delta$AUROC denotes the change in AUROC relative to the base model.}

\label{tab:chat_mia}
\end{table}

\paragraph{Experimental Findings.} As shown in Table~\ref{tab:rlvr_step20_delta}, RLVR induces substantial changes at the representation level. At the final layer, U shifts by $-0.53$ nats, against $-0.13$ under SFT. These shifts, however, show little member-specific bias: under SFT, deep layer $\mathrm{DiD}[D]$ is $+0.042$ and $\mathrm{DiD}[U]$ is $+0.182$, whereas under RLVR they are $+0.002$ and $+0.028$.

To examine this further at the behavioral level, we evaluate the model on the 64 member questions and 64 held-out non-member questions, sampling 32 completions per question at temperature 1.0 and averaging per-question accuracy across the 64 items. Before training, the two sets are essentially indistinguishable, with accuracies of 0.141 and 0.132. After RLVR training, member accuracy rises to 0.170 while non-member accuracy falls to 0.088, widening the gap from 0.009 to 0.082 (95\% CI [+0.008, +0.159]). Membership is thus directionally visible in behavior.

Why, then, are membership signals not readily detectable? A natural explanation is that the reward does not depend on $y^\star$, so training would not be expected to change its loss. Our results suggest otherwise: under RLVR, the loss on $y^\star$ increases by $0.24$ nats at step 10 and $0.56$ at step 20 for members. Non-members shift by the same amount, however, leaving a member/non-member gap below $0.01$ nats, compared with $-0.16$ nats after SFT on the same model and data. RLVR thus changes this loss without making it membership-specific, leaving likelihood-based attacks little signal to detect.

This is not a property of RL, as RL-based MIA methods show a clear member signal ($\mathrm{DiD}[U] = +0.185$ on AIME). What matters is how much the training signal depends on the member's own text. In SFT the signal comes from the cross-entropy loss, whose target is the member's own text. In RL-based MIA methods it comes from what the reward itself sees of the member, and under RLVR the reward sees only the final answer. We therefore read this case study as evidence that the scope of likelihood-based MIA is narrower than commonly assumed. \emph{\uline{Membership is hard to detect under low-supervision training. More precisely, likelihood-based attacks are most sensitive to training that directly affects the quantity they measure.}} 
\begin{table}[t]
\centering
\begin{tabular}{c|rrrrrr}
\toprule
& Layer 8 & Layer 12 & Layer 16 & Layer 20 & Layer 24 & Layer 28 \\
\midrule
$\Delta D$ & $-0.010$ & $-0.007$ & $-0.014$ & $-0.016$ & $-0.026$ & $-0.083$ \\
$\Delta U$ & $+0.079$ & $+0.139$ & $+0.217$ & $+0.195$ & $-0.001$ & $-0.525$ \\
\bottomrule
\end{tabular}
\caption{Changes in decodability ($\Delta D$) and readout ($\Delta U$) of RLVR relative to the base model.}
\label{tab:rlvr_step20_delta}
\end{table}

\subsection{Why not Steer?}

Analysis from Section~\ref{sec:dug_findings} shows that  SFT changes both decodable information and readout, whereas RL-based MIA methods act primarily on readout. This raises a practical question: can the same gain be obtained at inference time using steering vectors? However, this requires assuming that the target behavior is controlled by a single linear direction, an assumption that does not necessarily extend to readout amplification. We therefore test whether steering vectors can selectively strengthen MIA.

\paragraph{SFT.} We estimate the direction as the difference in mean residual-stream activations between member and nonmember texts at layer 24, which we choose as a representative mid-to-late layer. We define bootstrap consistency as the mean pairwise cosine similarity between directions estimated from bootstrap resamples. The SFT-induced direction is recovered stably (consistency 0.96). Single-layer injection yields no significant gain, with an AUROC of 0.56 and $p=0.07$, whereas sustained injection across layers 24–27 yields a modest gain, with AUROC increasing to 0.60 and $p=0.01$.

\paragraph{RL-based MIA.} We attempt the extraction for RL-based MIA methods as well, varying the subspace used to estimate the direction. We consider the full residual stream, the LoRA subspace, and the leading principal component of the training-induced shift $\Delta h$. The RL-induced direction is not recovered consistently, with bootstrap consistency ranging from 0.44 to 0.54, compared with 0.96 for its SFT counterpart. Although $\Delta h$ is nearly rank-one, with 78–95\% of its variance concentrated in one component (2-3.5\% for random directions), this dominant component is largely shared across members and nonmembers. This suggests that the RL-induced member-specific effect does not form a stable linear direction suitable for inference-time steering. Further details in App. \ref{app:steering}.

\subsection{Activation Patching}
\label{sec:patching}

To obtain causal evidence on where the membership-specific effect of RL-based MIA is formed, we perform activation patching. For a fixed input, we run $\theta_0$ and $\theta_a$ in parallel and swap the residual stream at the suffix positions of layer $\ell$ in both directions. 

Let $L_{\mathrm{out}}(y^\star_i)$ denote the model's cross-entropy on $y^\star_i$ under its full-vocabulary softmax. For any model $\theta$, we evaluate Eq.~\ref{eq:did} by replacing $\theta_a$ with $\theta$ and denote the result by $\mathrm{DiD}[X](\theta)$. We use $\mathrm{DiD}[X]$ here as shorthand for $\mathrm{DiD}[X](\theta_a)$. Let $\theta_{\mathrm{fwd},\ell}$ denote $\theta_0$ with its layer $\ell$ residual stream at the suffix positions replaced by that of $\theta_a$, and $\theta_{\mathrm{rev},\ell}$ denote $\theta_a$ with its layer $\ell$ residual stream replaced by that of $\theta_0$. We compute forward and reverse patching ratio as:

\begin{equation}
R_{\mathrm{fwd}}(\ell) = \frac{\mathrm{DiD}[-L_{\mathrm{out}}](\theta_{\mathrm{fwd},\ell})}{\mathrm{DiD}[-L_{\mathrm{out}}]}, \qquad
R_{\mathrm{rev}}(\ell) = \frac{\mathrm{DiD}[-L_{\mathrm{out}}](\theta_{\mathrm{rev},\ell})}{\mathrm{DiD}[-L_{\mathrm{out}}]}.
\end{equation}

Therefore, a positive $R_{\mathrm{fwd}}(\ell)-R_{\mathrm{rev}}(\ell)$ indicates that the effect is already present by layer $\ell$, while a negative value suggests that it emerges at later layers. We find that at $\ell=8$, $R_{\mathrm{fwd}} - R_{\mathrm{rev}}$ is significantly negative on both AIME ($[-1.410,-0.481]$) and Olympia Math ($[-1.201,-0.627]$), suggesting that the effect is not yet present. $\mathrm{DiD}[D]$ below this window is reported in Table~\ref{tab:did-d}. At $\ell=20$, it is significantly positive on Olympia Math ($[+0.307,+0.931]$), whereas the AIME estimate is inconclusive ($[-3.036,+0.594]$) because its denominator is close to zero. These results suggest that the effect emerges between layers 8 and 20 on Olympia Math, and after layer 8 on AIME.

\begin{table}[t]
\centering

\begin{tabular}{ccccc}
\toprule
Dataset & Layer & $\mathrm{DiD}[D]$ & 95\% CI & Crosses 0 \\
\midrule
AIME        & 4 & $-0.0013$ & $[-0.0146, +0.0124]$ & \checkmark \\
AIME        & 8 & $+0.0218$ & $[+0.0073, +0.0362]$ & \\
Olympia Math & 4 & $+0.0005$ & $[-0.0080, +0.0092]$ & \checkmark \\
Olympia Math & 8 & $-0.0045$ & $[-0.0122, +0.0029]$ & \checkmark \\
\bottomrule

\end{tabular}

%\caption{$\mathrm{DiD}[D]$ change ($\theta_a$ vs.\ $\theta_0$) below the patching window. }
\caption{$\mathrm{DiD}[D]$ ($\theta_a$ vs.\ $\theta_0$) at layers below the patching window. $\mathrm{DiD}[D]$ crosses zero in 3 of 4 cells, suggesting the window amplifies readout of an existing signal, though the AIME L8 cell indicates some contribution may be from storage.}
\label{tab:did-d}
\end{table}

\section{Discussion and Conclusion}

Our results suggest that membership signals depend not only on what is represented but also on how it is read out. SFT changes both decodable information and readout, whereas RL-based MIA methods act primarily on readout. Under our RLVR case study, behavioral changes do not necessarily translate into membership-specific likelihood signals when the reward is independent of the reference solution, leaving open whether similar patterns arise for non-likelihood-based MIA methods.

\newpage
\subsection*{AI use statement}

In this work, we used generative AI tools for polishing writing, specifically to improve readability and language; retrieval and discovery, specifically to locate sources for a small number of related works; research ideation or execution, specifically for feedback on experiments, coding and figure generation; and drafting sections of the paper, specifically for summarizing experimental settings and data in the appendix. We have not used generative AI tools for discovering research topics, identifying research gaps, formulating questions for literature surveys, summarizing existing literature, developing theoretical models or conceptual frameworks. Tasks of generating synthetic datasets, formulating mathematical claims, providing critical ingredients for proving mathematical claims, and assisting with writing proofs are not applicable to this work. 

We have reviewed all AI-assisted work. 
AI-edited code was verified and tested, and AI-assisted writing, descriptions of experimental settings, and extracted related-work content were manually reviewed for accuracy and consistency with the cited sources. We take responsibility for the final content of this work, including text, claims or artifacts produced with the aid of generative AI.

\bibliography{iclr2027_conference}

@misc{shi2024detectingpretrainingdatalarge,
      title={Detecting Pretraining Data from Large Language Models}, 
      author={Weijia Shi and Anirudh Ajith and Mengzhou Xia and Yangsibo Huang and Daogao Liu and Terra Blevins and Danqi Chen and Luke Zettlemoyer},
      year={2024},
      eprint={2310.16789},
      archivePrefix={arXiv},
      primaryClass={cs.CL},
      url={https://arxiv.org/abs/2310.16789}, 
}

@inproceedings{singh-etal-2024-aya,
    title = "Aya Dataset: An Open-Access Collection for Multilingual Instruction Tuning",
    author = {Singh, Shivalika  and
      Vargus, Freddie  and
      D{'}souza, Daniel  and
      Karlsson, B{\"o}rje F.  and
      Mahendiran, Abinaya  and
      Ko, Wei-Yin  and
      Shandilya, Herumb  and
      Patel, Jay  and
      Mataciunas, Deividas  and
      O{'}Mahony, Laura  and
      Zhang, Mike  and
      Hettiarachchi, Ramith  and
      Wilson, Joseph  and
      Machado, Marina  and
      Moura, Luisa  and
      Krzemi{\'n}ski, Dominik  and
      Fadaei, Hakimeh  and
      Ergun, Irem  and
      Okoh, Ifeoma  and
      Alaagib, Aisha  and
      Mudannayake, Oshan  and
      Alyafeai, Zaid  and
      Chien, Vu  and
      Ruder, Sebastian  and
      Guthikonda, Surya  and
      Alghamdi, Emad  and
      Gehrmann, Sebastian  and
      Muennighoff, Niklas  and
      Bartolo, Max  and
      Kreutzer, Julia  and
      {\"U}st{\"u}n, Ahmet  and
      Fadaee, Marzieh  and
      Hooker, Sara},
    editor = "Ku, Lun-Wei  and
      Martins, Andre  and
      Srikumar, Vivek",
    booktitle = "Proceedings of the 62nd Annual Meeting of the Association for Computational Linguistics (Volume 1: Long Papers)",
    month = aug,
    year = "2024",
    address = "Bangkok, Thailand",
    publisher = "Association for Computational Linguistics",
    url = "https://aclanthology.org/2024.acl-long.620/",
    doi = "10.18653/v1/2024.acl-long.620",
    pages = "11521--11567"
}

@misc{zhao2024wildchat1mchatgptinteraction,
      title={WildChat: 1M ChatGPT Interaction Logs in the Wild}, 
      author={Wenting Zhao and Xiang Ren and Jack Hessel and Claire Cardie and Yejin Choi and Yuntian Deng},
      year={2024},
      eprint={2405.01470},
      archivePrefix={arXiv},
      primaryClass={cs.CL},
      url={https://arxiv.org/abs/2405.01470}, 
}

@misc{hallinan2026surprisingeffectivenessmembershipinference,
      title={The Surprising Effectiveness of Membership Inference with Simple N-Gram Coverage}, 
      author={Skyler Hallinan and Jaehun Jung and Melanie Sclar and Ximing Lu and Abhilasha Ravichander and Sahana Ramnath and Yejin Choi and Sai Praneeth Karimireddy and Niloofar Mireshghallah and Xiang Ren},
      year={2026},
      eprint={2508.09603},
      archivePrefix={arXiv},
      primaryClass={cs.CL},
      url={https://arxiv.org/abs/2508.09603}, 
}

@misc{yin2026learningdetectlanguagemodel,
      title={Learning to Detect Language Model Training Data via Active Reconstruction}, 
      author={Junjie Oscar Yin and John X. Morris and Vitaly Shmatikov and Sewon Min and Hannaneh Hajishirzi},
      year={2026},
      eprint={2602.19020},
      archivePrefix={arXiv},
      primaryClass={cs.LG},
      url={https://arxiv.org/abs/2602.19020}, 
}

@misc{das2025blindbaselinesbeatmembership,
      title={Blind Baselines Beat Membership Inference Attacks for Foundation Models}, 
      author={Debeshee Das and Jie Zhang and Florian Tramèr},
      year={2025},
      eprint={2406.16201},
      archivePrefix={arXiv},
      primaryClass={cs.CR},
      url={https://arxiv.org/abs/2406.16201}, 
}

@misc{belrose2025elicitinglatentpredictionstransformers,
      title={Eliciting Latent Predictions from Transformers with the Tuned Lens}, 
      author={Nora Belrose and Igor Ostrovsky and Lev McKinney and Zach Furman and Logan Smith and Danny Halawi and Stella Biderman and Jacob Steinhardt},
      year={2025},
      eprint={2303.08112},
      archivePrefix={arXiv},
      primaryClass={cs.LG},
      url={https://arxiv.org/abs/2303.08112}, 
}

@inproceedings{
skean2025layer,
title={Layer by Layer: Uncovering Hidden Representations in Language Models},
author={Oscar Skean and Md Rifat Arefin and Dan Zhao and Niket Nikul Patel and Jalal Naghiyev and Yann LeCun and Ravid Shwartz-Ziv},
booktitle={Forty-second International Conference on Machine Learning},
year={2025},
url={https://openreview.net/forum?id=WGXb7UdvTX}
}

@misc{shokri2017membershipinferenceattacksmachine,
      title={Membership Inference Attacks against Machine Learning Models}, 
      author={Reza Shokri and Marco Stronati and Congzheng Song and Vitaly Shmatikov},
      year={2017},
      eprint={1610.05820},
      archivePrefix={arXiv},
      primaryClass={cs.CR},
      url={https://arxiv.org/abs/1610.05820}, 
}

@misc{rl-data-contamination,
      title={Detecting Data Contamination from Reinforcement Learning Post-training for Large Language Models}, 
      author={Yongding Tao and Tian Wang and Yihong Dong and Huanyu Liu and Kechi Zhang and Xiaolong Hu and Ge Li},
      year={2025},
      eprint={2510.09259},
      archivePrefix={arXiv},
      primaryClass={cs.CL},
      url={https://arxiv.org/abs/2510.09259}, 
}

@misc{zhang2026detectingrlvrtrainingdata,
      title={Detecting RLVR Training Data via Structural Convergence of Reasoning}, 
      author={Hongbo Zhang and Yue Yang and Jianhao Yan and Guangsheng Bao and Yue Zhang and Yue Zhang},
      year={2026},
      eprint={2602.11792},
      archivePrefix={arXiv},
      primaryClass={cs.AI},
      url={https://arxiv.org/abs/2602.11792}, 
}

@misc{fu2024miatuneradaptinglargelanguage,
      title={MIA-Tuner: Adapting Large Language Models as Pre-training Text Detector}, 
      author={Wenjie Fu and Huandong Wang and Chen Gao and Guanghua Liu and Yong Li and Tao Jiang},
      year={2024},
      eprint={2408.08661},
      archivePrefix={arXiv},
      primaryClass={cs.CL},
      url={https://arxiv.org/abs/2408.08661}, 
}

@misc{liu2026auditingdatamembershipreinforcement,
      title={Auditing Data Membership in Reinforcement Learning With Verifiable Rewards}, 
      author={Yule Liu and Heyi Zhang and Jinyi Zheng and Zhen Sun and Zifan Peng and Jiaheng Wei and Tianshuo Cong and Yilong Yang and Xinlei He},
      year={2026},
      eprint={2511.14045},
      archivePrefix={arXiv},
      primaryClass={cs.CR},
      url={https://arxiv.org/abs/2511.14045}, 
}

@misc{gwak2026laralayerwiserepresentationanalysis,
      title={LaRA: Layer-wise Representation Analysis for Detecting Data Contamination in RL Post-Training}, 
      author={Minju Gwak and Minseo Kwak and Dongseok Lee and Guijin Son and Alan Ritter and Jaehyung Kim},
      year={2026},
      eprint={2605.29888},
      archivePrefix={arXiv},
      primaryClass={cs.LG},
      url={https://arxiv.org/abs/2605.29888}, 
}

@misc{tao2025tokenlevelinformiastrongermembership,
      title={(Token-Level) InfoRMIA: Stronger Membership Inference and Memorization Assessment for LLMs}, 
      author={Jiashu Tao and Reza Shokri},
      year={2025},
      eprint={2510.05582},
      archivePrefix={arXiv},
      primaryClass={cs.LG},
      url={https://arxiv.org/abs/2510.05582}, 
}

@misc{yan2026spuriousrewardsparadoxmechanistically,
      title={Spurious Rewards Paradox: Mechanistically Understanding How RLVR Activates Memorization Shortcuts in LLMs}, 
      author={Lecheng Yan and Ruizhe Li and Guanhua Chen and Qing Li and Jiahui Geng and Wenxi Li and Longyue Wang and Chenyang Lyu},
      year={2026},
      eprint={2601.11061},
      archivePrefix={arXiv},
      primaryClass={cs.LG},
      url={https://arxiv.org/abs/2601.11061}, 
}

@inproceedings{xie-etal-2025-memorization,
    title = "On Memorization of Large Language Models in Logical Reasoning",
    author = "Xie, Chulin  and
      Huang, Yangsibo  and
      Zhang, Chiyuan  and
      Yu, Da  and
      Chen, Xinyun  and
      Lin, Bill Yuchen  and
      Li, Bo  and
      Ghazi, Badih  and
      Kumar, Ravi",
    editor = "Inui, Kentaro  and
      Sakti, Sakriani  and
      Wang, Haofen  and
      Wong, Derek F.  and
      Bhattacharyya, Pushpak  and
      Banerjee, Biplab  and
      Ekbal, Asif  and
      Chakraborty, Tanmoy  and
      Singh, Dhirendra Pratap",
    booktitle = "Proceedings of the 14th International Joint Conference on Natural Language Processing and the 4th Conference of the Asia-Pacific Chapter of the Association for Computational Linguistics",
    month = dec,
    year = "2025",
    address = "Mumbai, India",
    publisher = "The Asian Federation of Natural Language Processing and The Association for Computational Linguistics",
    url = "https://aclanthology.org/2025.ijcnlp-long.148/",
    doi = "10.18653/v1/2025.ijcnlp-long.148",
    pages = "2742--2785",
    ISBN = "979-8-89176-298-5"
}

@misc{carlini2021extractingtrainingdatalarge,
      title={Extracting Training Data from Large Language Models}, 
      author={Nicholas Carlini and Florian Tramer and Eric Wallace and Matthew Jagielski and Ariel Herbert-Voss and Katherine Lee and Adam Roberts and Tom Brown and Dawn Song and Ulfar Erlingsson and Alina Oprea and Colin Raffel},
      year={2021},
      eprint={2012.07805},
      archivePrefix={arXiv},
      primaryClass={cs.CR},
      url={https://arxiv.org/abs/2012.07805}, 
}

@misc{duan2024membershipinferenceattackswork,
      title={Do Membership Inference Attacks Work on Large Language Models?}, 
      author={Michael Duan and Anshuman Suri and Niloofar Mireshghallah and Sewon Min and Weijia Shi and Luke Zettlemoyer and Yulia Tsvetkov and Yejin Choi and David Evans and Hannaneh Hajishirzi},
      year={2024},
      eprint={2402.07841},
      archivePrefix={arXiv},
      primaryClass={cs.CL},
      url={https://arxiv.org/abs/2402.07841}, 
}

@inproceedings{lee-etal-2022-deduplicating,
    title = "Deduplicating Training Data Makes Language Models Better",
    author = "Lee, Katherine  and
      Ippolito, Daphne  and
      Nystrom, Andrew  and
      Zhang, Chiyuan  and
      Eck, Douglas  and
      Callison-Burch, Chris  and
      Carlini, Nicholas",
    editor = "Muresan, Smaranda  and
      Nakov, Preslav  and
      Villavicencio, Aline",
    booktitle = "Proceedings of the 60th Annual Meeting of the Association for Computational Linguistics (Volume 1: Long Papers)",
    month = may,
    year = "2022",
    address = "Dublin, Ireland",
    publisher = "Association for Computational Linguistics",
    url = "https://aclanthology.org/2022.acl-long.577/",
    doi = "10.18653/v1/2022.acl-long.577",
    pages = "8424--8445"
}

@misc{chen2026comparativeanalysisllmmemorization,
      title={A Comparative Analysis of LLM Memorization at Statistical and Internal Levels: Cross-Model Commonalities and Model-Specific Signatures}, 
      author={Bowen Chen and Namgi Han and Yusuke Miyao},
      year={2026},
      eprint={2603.21658},
      archivePrefix={arXiv},
      primaryClass={cs.CL},
      url={https://arxiv.org/abs/2603.21658}, 
}

@misc{stoehr2024localizingparagraphmemorizationlanguage,
      title={Localizing Paragraph Memorization in Language Models}, 
      author={Niklas Stoehr and Mitchell Gordon and Chiyuan Zhang and Owen Lewis},
      year={2024},
      eprint={2403.19851},
      archivePrefix={arXiv},
      primaryClass={cs.CL},
      url={https://arxiv.org/abs/2403.19851}, 
}

@inproceedings{prabhakar-etal-2024-deciphering,
    title = "Deciphering the Factors Influencing the Efficacy of Chain-of-Thought: Probability, Memorization, and Noisy Reasoning",
    author = "Prabhakar, Akshara  and
      Griffiths, Thomas L.  and
      McCoy, R. Thomas",
    editor = "Al-Onaizan, Yaser  and
      Bansal, Mohit  and
      Chen, Yun-Nung",
    booktitle = "Findings of the Association for Computational Linguistics: EMNLP 2024",
    month = nov,
    year = "2024",
    address = "Miami, Florida, USA",
    publisher = "Association for Computational Linguistics",
    url = "https://aclanthology.org/2024.findings-emnlp.212/",
    doi = "10.18653/v1/2024.findings-emnlp.212",
    pages = "3710--3724"
}

@misc{stolfo2025improvinginstructionfollowinglanguagemodels,
      title={Improving Instruction-Following in Language Models through Activation Steering}, 
      author={Alessandro Stolfo and Vidhisha Balachandran and Safoora Yousefi and Eric Horvitz and Besmira Nushi},
      year={2025},
      eprint={2410.12877},
      archivePrefix={arXiv},
      primaryClass={cs.CL},
      url={https://arxiv.org/abs/2410.12877}, 
}

@article{gurnee2026verbalizable,
  author={Gurnee, Wes and Sofroniew, Nicholas and Pearce, Adam and Piotrowski, Mateusz and Kauvar, Isaac and Chen, Runjin and Soligo, Anna and Bogdan, Paul and Ong, Euan and Wang, Rowan and Thompson, Ben and Abrahams, David and Kantamneni, Subhash and Ameisen, Emmanuel and Batson, Joshua and Lindsey, Jack},
  title={Verbalizable Representations Form a Global Workspace in Language Models},
  journal={Transformer Circuits Thread},
  year={2026},
  url={https://transformer-circuits.pub/2026/workspace/index.html}
}

@misc{carlini2023quantifyingmemorizationneurallanguage,
      title={Quantifying Memorization Across Neural Language Models}, 
      author={Nicholas Carlini and Daphne Ippolito and Matthew Jagielski and Katherine Lee and Florian Tramer and Chiyuan Zhang},
      year={2023},
      eprint={2202.07646},
      archivePrefix={arXiv},
      primaryClass={cs.LG},
      url={https://arxiv.org/abs/2202.07646}, 
}

@misc{shao2024deepseekmathpushinglimitsmathematical,
      title={DeepSeekMath: Pushing the Limits of Mathematical Reasoning in Open Language Models}, 
      author={Zhihong Shao and Peiyi Wang and Qihao Zhu and Runxin Xu and Junxiao Song and Xiao Bi and Haowei Zhang and Mingchuan Zhang and Y. K. Li and Y. Wu and Daya Guo},
      year={2024},
      eprint={2402.03300},
      archivePrefix={arXiv},
      primaryClass={cs.CL},
      url={https://arxiv.org/abs/2402.03300}, 
}

@inproceedings{makhija-etal-2026-neural,
    title = "Neural Breadcrumbs: Membership Inference Attacks on {LLM}s Through Hidden State and Attention Pattern Analysis",
    author = "Makhija, Disha  and
      Arivazhagan, Manoj Ghuhan  and
      Kumar, Vinayshekhar Bannihatti  and
      Gangadharaiah, Rashmi",
    editor = "Demberg, Vera  and
      Inui, Kentaro  and
      Marquez, Llu{\'i}s",
    booktitle = "Proceedings of the 19th Conference of the {E}uropean Chapter of the {A}ssociation for {C}omputational {L}inguistics (Volume 1: Long Papers)",
    month = mar,
    year = "2026",
    address = "Rabat, Morocco",
    publisher = "Association for Computational Linguistics",
    url = "https://aclanthology.org/2026.eacl-long.262/",
    doi = "10.18653/v1/2026.eacl-long.262",
    pages = "5608--5620",
    ISBN = "979-8-89176-380-7"
}

@inproceedings{
wang2025generalization,
title={Generalization v.s. Memorization: Tracing Language Models{\textquoteright} Capabilities Back to Pretraining Data},
author={Xinyi Wang and Antonis Antoniades and Yanai Elazar and Alfonso Amayuelas and Alon Albalak and Kexun Zhang and William Yang Wang},
booktitle={The Thirteenth International Conference on Learning Representations},
year={2025},
url={https://openreview.net/forum?id=IQxBDLmVpT}
}

@misc{ivison2023camelschangingclimateenhancing,
      title={Camels in a Changing Climate: Enhancing LM Adaptation with Tulu 2}, 
      author={Hamish Ivison and Yizhong Wang and Valentina Pyatkin and Nathan Lambert and Matthew Peters and Pradeep Dasigi and Joel Jang and David Wadden and Noah A. Smith and Iz Beltagy and Hannaneh Hajishirzi},
      year={2023},
      eprint={2311.10702},
      archivePrefix={arXiv},
      primaryClass={cs.CL},
      url={https://arxiv.org/abs/2311.10702}, 
}

@misc{yang2024qwen2technicalreport,
      title={Qwen2 Technical Report}, 
      author={An Yang and Baosong Yang and Binyuan Hui and Bo Zheng and Bowen Yu and Chang Zhou and Chengpeng Li and Chengyuan Li and Dayiheng Liu and Fei Huang and Guanting Dong and Haoran Wei and Huan Lin and Jialong Tang and Jialin Wang and Jian Yang and Jianhong Tu and Jianwei Zhang and Jianxin Ma and Jianxin Yang and Jin Xu and Jingren Zhou and Jinze Bai and Jinzheng He and Junyang Lin and Kai Dang and Keming Lu and Keqin Chen and Kexin Yang and Mei Li and Mingfeng Xue and Na Ni and Pei Zhang and Peng Wang and Ru Peng and Rui Men and Ruize Gao and Runji Lin and Shijie Wang and Shuai Bai and Sinan Tan and Tianhang Zhu and Tianhao Li and Tianyu Liu and Wenbin Ge and Xiaodong Deng and Xiaohuan Zhou and Xingzhang Ren and Xinyu Zhang and Xipin Wei and Xuancheng Ren and Xuejing Liu and Yang Fan and Yang Yao and Yichang Zhang and Yu Wan and Yunfei Chu and Yuqiong Liu and Zeyu Cui and Zhenru Zhang and Zhifang Guo and Zhihao Fan},
      year={2024},
      eprint={2407.10671},
      archivePrefix={arXiv},
      primaryClass={cs.CL},
      url={https://arxiv.org/abs/2407.10671}, 
}

@misc{ai-mo2025olympiad,
  author       = {{AI-MO}},
  title        = {AI-MO Olympiad Reference Dataset},
  year         = {2025},
  howpublished = {Hugging Face Dataset},
  url          = {https://huggingface.co/datasets/AI-MO/olympiads-ref}
}

@inproceedings{
lightman2024lets,
title={Let's Verify Step by Step},
author={Hunter Lightman and Vineet Kosaraju and Yuri Burda and Harrison Edwards and Bowen Baker and Teddy Lee and Jan Leike and John Schulman and Ilya Sutskever and Karl Cobbe},
booktitle={The Twelfth International Conference on Learning Representations},
year={2024},
url={https://openreview.net/forum?id=v8L0pN6EOi}
}

@misc{magar2022datacontaminationmemorizationexploitation,
      title={Data Contamination: From Memorization to Exploitation}, 
      author={Inbal Magar and Roy Schwartz},
      year={2022},
      eprint={2203.08242},
      archivePrefix={arXiv},
      primaryClass={cs.CL},
      url={https://arxiv.org/abs/2203.08242}, 
}

@inproceedings{
oren2024proving,
title={Proving Test Set Contamination in Black-Box Language Models},
author={Yonatan Oren and Nicole Meister and Niladri S. Chatterji and Faisal Ladhak and Tatsunori Hashimoto},
booktitle={The Twelfth International Conference on Learning Representations},
year={2024},
url={https://openreview.net/forum?id=KS8mIvetg2}
}

@misc{jiang2024investigatingdatacontaminationpretraining,
      title={Investigating Data Contamination for Pre-training Language Models}, 
      author={Minhao Jiang and Ken Ziyu Liu and Ming Zhong and Rylan Schaeffer and Siru Ouyang and Jiawei Han and Sanmi Koyejo},
      year={2024},
      eprint={2401.06059},
      archivePrefix={arXiv},
      primaryClass={cs.CL},
      url={https://arxiv.org/abs/2401.06059}, 
}

@misc{wu2025reasoningmemorizationunreliableresults,
      title={Reasoning or Memorization? Unreliable Results of Reinforcement Learning Due to Data Contamination}, 
      author={Mingqi Wu and Zhihao Zhang and Qiaole Dong and Zhiheng Xi and Jun Zhao and Senjie Jin and Xiaoran Fan and Yuhao Zhou and Huijie Lv and Ming Zhang and Yanwei Fu and Qin Liu and Songyang Zhang and Qi Zhang},
      year={2025},
      eprint={2507.10532},
      archivePrefix={arXiv},
      primaryClass={cs.LG},
      url={https://arxiv.org/abs/2507.10532}, 
}
\bibliographystyle{iclr2027_conference}

\newpage
\appendix
%\section{Preliminary Experiments}
%\label{app:prelim}
%\subsection{Settings}

\section{DUG Experiments}

\label{app:DUG_exp}
\subsection{Dataset and Model}
\label{app:dug_dataset}

\paragraph{Dataset.} For the controlled-contamination setting, we use two math datasets, AIME and Olympia Math. For AIME, we collect problems and solutions from 2021–2025. Since each year contains only 30 problems, this gives 120 member examples and 30 non-member examples. For each seed, we draw 32 examples, with 16 from each group. For Olympia Math, we divide the problems into member and non-member groups using a random partition, and draw 64 examples per seed, with 32 from each group. For WikiMIA, members and non-members are defined by a temporal cutoff. From each group, we draw 64 examples for evaluation. All examples used as members are randomly selected from their respective member pools.

Our evaluation sets are intentionally small, reflecting a light-contamination setting we target, where only a limited number of benchmark items leak into training. This mirrors how contamination typically arises in practice. It also follows controlled-contamination protocols in prior work, which inject a limited number of test examples into continued pre-training or fine-tuning to study their detectability \citep{shi2024detectingpretrainingdatalarge, jiang2024investigatingdatacontaminationpretraining, oren2024proving}. Heavier contamination would shift the focus of our analysis.

\paragraph{Choice of Models.} A controlled study requires us to know whether exposure to the evaluation data comes from our training \citep{duan2024membershipinferenceattackswork}. This is difficult to establish for recent models, whose training corpora are not fully disclosed and may already contain widely circulated benchmarks \citep{wu2025reasoningmemorizationunreliableresults}. We therefore use earlier base models and measure membership effects relative to the base model, so that prior exposure shared by members and non-members is accounted for. 

\subsection{Explanation on RL-based MIA Methods}

\label{app:explanation_RL}

Throughout, $\theta_0$ denotes the base model before any of our training. RL-based MIA methods are optimized on the same members and non-members as the other settings. In the post-training setting (AIME and Olympia Math), they are initialized from the SFT model after one epoch, so that membership is introduced before the attack; in the pre-training setting (WikiMIA-2024-Hard), they are initialized from $\theta_0$. All DiD values are computed relative to $\theta_0$. Since DiD is linear in the trained model's scores, the difference between the RL-based MIA methods and SFT (1 ep) curves equals the DiD of RL-based MIA methods relative to their initialization, i.e., the effect of RL alone.

\subsection{Framework Details}

\paragraph{Output vocabulary.}
For probing, predicting over the full vocabulary would make the probe's output head dominate its
parameter count, especially for Qwen2-7B ($|\mathcal{V}| = 152{,}064$). We therefore
restrict the output space to the tokens observed in each dataset's suffix pool plus a
single \texttt{UNK} class: $1{,}328$ classes for AIME and $2{,}630$ for Olympia Math
(both on tulu-2-7b, $|\mathcal{V}| = 32{,}000$), and $6{,}005$ for WikiMIA-2024-Hard.
The same restriction is applied to the null baseline and to the logit lens.

Hence the probe, the logit lens, and the null
baseline are all evaluated on the same output space, and $D(\ell)$, $U(\ell)$, and
$G(\ell)$ are directly comparable.

\paragraph{Probe capacity.} At each layer, we fit a linear probe. For AIME and Olympia Math, we use full rank probe. For WikiMIA-2024-Hard, a full-rank probe would be heavily over-parameterized relative to the available training samples, so we cap the rank at $r=256$. This cap is held fixed across all layers within this dataset, so the layerwise profiles remain internally comparable.

\paragraph{Controls and inference.}

Probes are fit using 5-fold GroupKFold cross-validation, with problems as groups. For each layer, the weight decay is selected from ${10^{-4},10^{-3},10^{-2},10^{-1}}$ using a shuffle-label control, choosing the smallest value for which the shuffled-target probe yields $|D_{\mathrm{shuffle}}| \leq 0.05H_{\mathrm{pool}}$. This threshold corresponds to $0.270$, $0.295$, and $0.338$ nats for AIME, Olympia Math, and WikiMIA, respectively. Confidence intervals and DiD tests are computed from $1{,}000$ problem-level bootstrap samples, with members and non-members resampled independently.

\subsection{Binary Membership Detection}
\label{app:binary-probe}
Following our preliminary experiments as in Sec.~\ref{sec:prelim}, we also examine whether a binary probe can detect membership-related changes in the trained models. Specifically, we fit a logistic regression on mean- and last-token residual-stream features at each layer, after PCA ($k=30$) and residualization on suffix length, using Olympia Math member/non-member splits of Sec.~\ref{sec:design_dug_sec5}, and compare against a shuffled-label control and the loss-based MIA AUROC of the same models.

Across layers, probe AUROC on $\theta_0$ ranges from 0.29 to 0.55, while the shuffled-label control remains near 0.50. For SFT, the probe yields AUROC of 0.30–0.55, with no clear improvement over $\theta_0$. At this scale, SFT produces a modest increase in loss-based MIA performance (AUROC 0.61), while the token-level measures of DUG reveal membership-specific changes (e.g., $\mathrm{DiD}[U]$ at late layers, as in Figure~\ref{fig:DiD}). These results suggest that per-example AUROC may be less sensitive to membership effects at this sample scale, motivating our use of token-level measures. We therefore do not use the binary probe to draw conclusions about RL-based MIA methods.

\subsection{Further Results}

\subsubsection{Results on WikiMIA-2024-Hard.} 

\label{app:wikimia}

\begin{figure}[htp]
    \centering
    \includegraphics[width=1\linewidth]{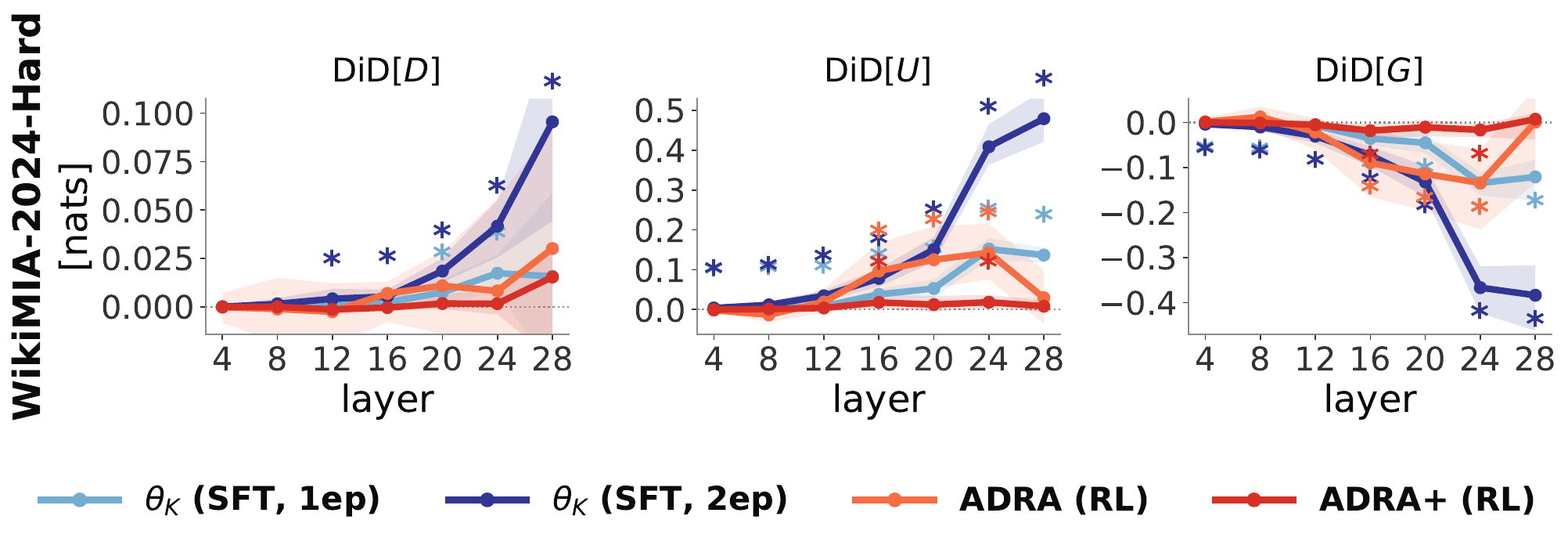}
    \caption{Diff-in-differences of membership-related changes in decodability ($D$), readout ($U$), and their gap ($G$) across layers.}
    \label{fig:Wiki_DiD}
\end{figure}

Figure~\ref{fig:Wiki_DiD} reports the difference-in-difference estimates on WikiMIA-2024-Hard. SFT reproduces the main-text pattern, in which both $\mathrm{DiD}[D]$ and $\mathrm{DiD}[U]$ grow with epochs and are significant at most layers. RL-based MIA methods show the same readout-specific effect, with the asterisks in the figure indicating statistical significance, though in a weaker form.

We attribute the weaker RL effects to two factors. First, RL here starts from pre-training exposure, where the membership-specific signal appears relatively difficult to further amplify. Our preliminary results likewise suggest that signals established during the pre-training stage are less amenable to amplification by subsequent training. Second, ADRA+ gates the ground-truth suffix with a Min-K\%++ prior whose standalone AUROC is 50.6 on WikiMIA-2024-Hard, compared with 78.3 on AIME and 60.2 on Olympia Math \citep{yin2026learningdetectlanguagemodel}. At chance level, the gate is unrelated to membership, so no
member-specific effect accumulates. ADRA+ effect sizes follow the same ordering as the prior AUROC, though three datasets cannot establish this causally.

\subsubsection{DiD relative to the SFT 1ep.}

\label{app:sft1ep}

In the post-training setting, RL-based MIA methods are initialized from the SFT (1 ep) model, so their DiD relative to $\theta_0$ also includes the effect of SFT. We already report the DiD for SFT (1 ep) in the main text for reference. For completeness, we additionally compute Eq.~\ref{eq:did} using the SFT (1 ep) model in place of $\theta_0$ on AIME and Olympia Math.

\begin{figure}[htp]
    \centering
    \includegraphics[width=1\linewidth]{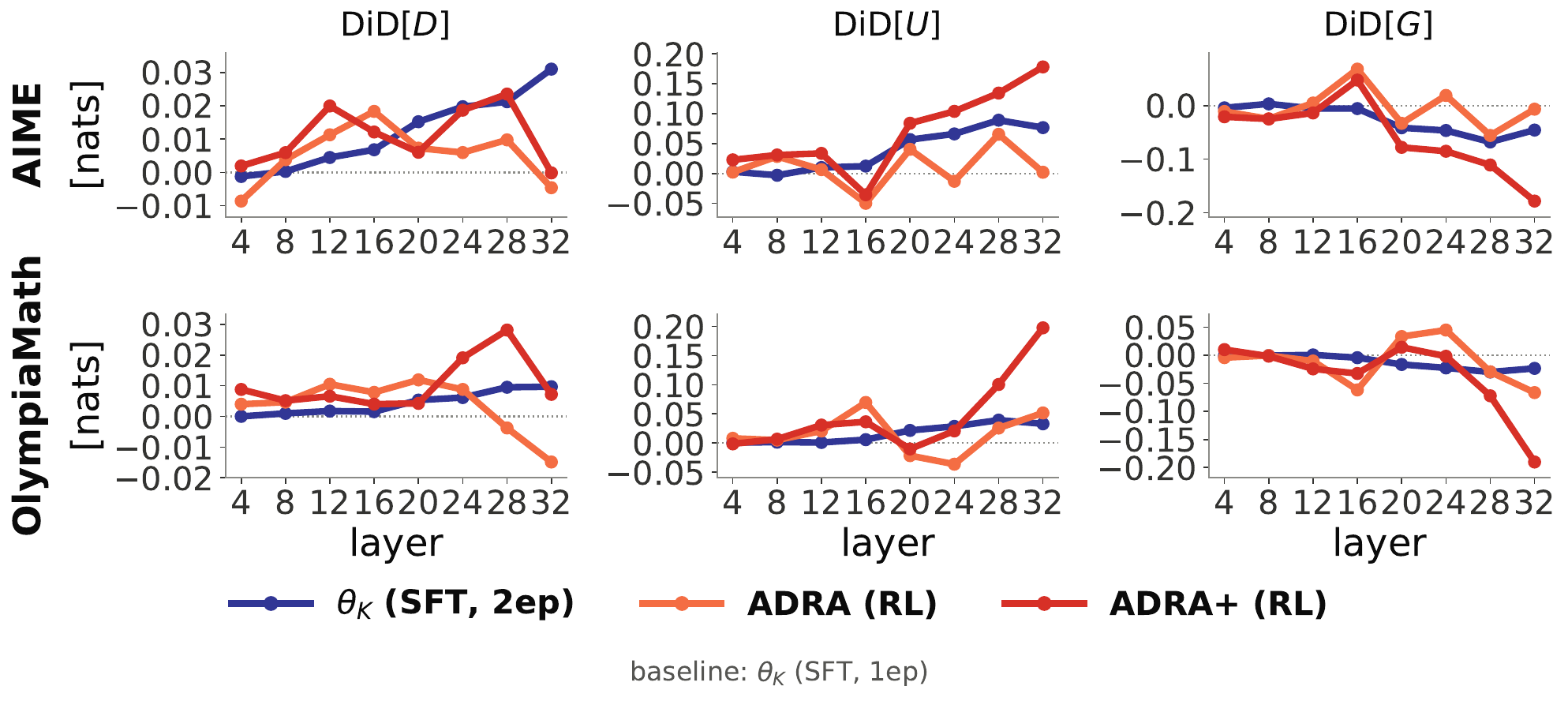}
    \caption{Further results on membership-specific changes. DiD scores for D and U using SFT (1 ep) as the reference model.}
    \label{fig:did_sft_ep1}
\end{figure}

As shown in Figure~\ref{fig:did_sft_ep1}, $\mathrm{DiD}[U]$ remains positive at mid-to-late layers, reaching $+0.18$ nats on AIME and $+0.20$ nats on Olympia Math, whereas $\mathrm{DiD}[D]$ stays very close to zero. This is consistent with Section~\ref{sec:dug_findings} and supports our conclusion that RL-based MIA methods amplify membership-specific effects primarily through readout rather than through decodable information.

\section{Why Not a Tuned Lens?}
\label{app:tuned_lens}

We also implemented a tuned lens~\citep{belrose2025elicitinglatentpredictionstransformers}, which learns an affine translator $(A_\ell, c_\ell)$ per layer by minimizing the KL divergence to the final-layer distribution $p_{\mathrm{final}}$, using the same folds as the probe. We did not adopt it for two reasons.

\paragraph{The translator absorbs later computation.} Because $A_\ell$ is trained to reproduce $p_{\mathrm{final}}$, the tuned lens measures how well layer $\ell$ can be mapped onto the model's final prediction, rather than what the model reads out at layer $\ell$. Empirically, $U_{\mathrm{tuned}}$ exceeds $D$ at all layers, yielding a negative $G$ throughout.

\paragraph{Trainable readouts can introduce membership differences.}
The bias $c_\ell$ learns a fold-level constant related to the mean of $p_{\mathrm{final}}$. Because members and non-members differ in final-layer confidence, this constant can induce a member/non-member difference in $U_{\mathrm{tuned}}$. The difference is largest at layer 0 (0.196 nats), where the representation contains little membership information, and decreases to 0.005 nats at the final layer, close to the logit-lens value of 0.004. This suggests that trainable readouts may introduce membership differences beyond those reflected in the model’s own output head, motivating the use of the logit lens for $U$.

\section{RLVR Case Study}

\subsection{Experimental Settings}
\label{app:RLVR_setting}
Members and nonmembers are both drawn from MATH-500 problems with $0 < \mathrm{pass@32} < 1$, ensuring GRPO's group-normalized advantage is nonzero \citep{shao2024deepseekmathpushinglimitsmathematical}. The 64 members are mixed into 576 problems from the training set, yielding 640 prompts; the nonmembers never enter training at any stage. Each record retains only the problem statement and the final answer. With 64 prompts per optimizer step, ten steps form one pool epoch in which every member appears exactly once, so RLVR at steps 10 and 20 is exposure-matched to SFT at 1 and 2 epochs. Training ran until policy entropy fell below half its step-1 value. 

\subsection{Supplementary Training Validity}
\label{app:RLVR_valid}
Mean pool accuracy rises from 54.9\% in epoch 1 to 71.8\% in epoch 2, with per-step accuracy increasing from $56.5\%$ to 83.4\%. Mean reward rises correspondingly from 0.613 to 0.884, while reward variance and the fraction of groups with non-zero advantage decrease, consistent with the policy approaching the task ceiling.

\subsection{Future Directions}
\label{app:RLVR_future}
Our analysis of the RLVR case study focuses on likelihood-based signals. Evaluating non-likelihood-based detection methods is orthogonal to our question and remains a direction for future work, along with broader training objectives.

\section{Additional Details on Activation Steering}
\label{app:steering}

\paragraph{SFT.} For bootstrap consistency, we use 100 resamples of the 32 construction pairs. The resulting directions have a mean pairwise cosine similarity of 0.96.

To examine the limited effect of single-layer injection, we measure how much of the injected component remains at the final layer. The corresponding measure is 0.05, compared with 0.75 for random directions, indicating that the injected component is largely weakened through the remaining layers.

For sustained injection, we evaluate the intervention on 183 held-out pairs. Reversing the injection direction yields chance-level performance.

\end{document}